\documentclass[conference]{IEEEtran}
\usepackage[noadjust]{cite}

\usepackage{graphicx}
\usepackage{amsmath}
\usepackage{color,soul}
\usepackage{multirow}
\usepackage{csquotes}
\usepackage[ruled,lined,boxed,linesnumbered]{algorithm2e}
\usepackage{authblk}
\author[1]{\large Ji Li\thanks{jli724@usc.edu}}
\author[1]{Zihao Yuan\thanks{zihaoyua@usc.edu}}
\author[2]{Zhe Li\thanks{zli89@syr.edu}}
\author[2]{Caiwen Ding\thanks{cading@syr.edu}}
\author[2]{Ao Ren\thanks{aren@syr.edu}}
\author[2]{Qinru Qiu\thanks{qiqiu@syr.edu}}
\author[1,3]{Jeffrey Draper\thanks{draper@isi.edu}}
\author[2]{Yanzhi Wang\thanks{ywang393@syr.edu}}
\affil[1]{\normalsize Department of Electrical Engineering, University of Southern California, Los Angeles, CA, USA}
\affil[2]{College of Engineering and Computer Science, Syracuse University, Syracuse, NY, USA}
\affil[3]{Information Sciences Institute, University of Southern California, Marina Del Rey, CA, USA}
\affil[ ]{\textit {\{jli724,zihaoyua\}@usc.edu, \{zli89,cading,aren,qiqiu\}@syr.edu, draper@isi.edu, ywang393@syr.edu}
}

\begin{document}
\bstctlcite{IEEEexample:BSTcontrol}


\title{ Hardware-Driven Nonlinear Activation for Stochastic Computing Based Deep Convolutional Neural Networks}

%
%
%
\maketitle

\begin{abstract}
Recently, Deep Convolutional Neural Networks (DCNNs) have made unprecedented progress, achieving the accuracy close to, or even better than human-level perception in various tasks. 
There is a timely need to map the latest software DCNNs to application-specific hardware, in order to achieve orders of magnitude improvement in performance, energy efficiency and compactness. 
Stochastic Computing (SC), as a low-cost alternative to the conventional binary computing paradigm, has the potential to enable massively parallel and highly scalable hardware implementation of DCNNs. 
One major challenge in SC based DCNNs is designing accurate nonlinear activation functions, which have a significant impact on the network-level accuracy but cannot be implemented accurately by existing SC computing blocks. 
In this paper, we design and optimize SC based neurons, and we propose highly accurate activation designs for the three most frequently used activation functions in software DCNNs, i.e, hyperbolic tangent, logistic, and rectified linear units. 
Experimental results on LeNet-5 using MNIST dataset demonstrate that compared with a binary ASIC hardware DCNN, the DCNN with the proposed SC neurons can achieve up to 61X, 151X, and 2X improvement in terms of area, power, and energy, respectively, at the cost of small precision degradation. 
In addition, the SC approach achieves up to 21X and 41X of the area, 41X and 72X of the power, and 198200X and 96443X of the energy, compared with CPU and GPU approaches, respectively, while the error is increased by less than 3.07\%.
ReLU activation is suggested for future SC based DCNNs considering its superior performance under a small bit stream length.  

\end{abstract}


\section{Introduction}

Deep learning has achieved unprecedented success in solving problems that have resisted the best attempts of the artificial intelligence community for many years \cite{bengio2009learning,lecun2015deep}.
Artificial neural networks with deep learning can automatically extract a hierarchy of representations by composing nonlinear modules, which transform the representations at one level (starting from raw input, e.g., pixel values of an image) into representations at a more abstract and more invariant level \cite{lecun2015deep}. 
Arbitrarily complex functions can be approximated in a sufficiently large neural network using \textit{nonlinear activation functions} \cite{cho2010large}. However, in practice the sizes of networks are finite, and the choice of nonlinearity affects both the learning dynamics and the network\rq s expressive power \cite{agostinelli2014learning}.


Recently, the Deep Convolutional Neural Network (DCNN) has achieved breakthroughs in many fundamental applications, such as image/video classification \cite{simonyan2014very,karpathy2014large} and visual/text recognition \cite{donahue2014decaf,7850026}.
DCNN is now recognized as the dominant method for almost all recognition and detection tasks and surpasses human performance on certain tasks \cite{lecun2015deep}. 
Since the deep layered structures require a large amount of computation resources, from a practical standpoint, large-scale DCNNs are mainly implemented in high performance server clusters \cite{ardakani2015vlsi}.  
The huge power/energy consumptions of software DCNNs prevent their widespread deployment in wearable and Internet of Things (IoT) devices, which emerge with repercussions across the industry spectrum \cite{kim2016dynamic}. 
Hence, there is a timely need to map the latest DCNNs to application-specific hardware, in order to achieve orders of magnitude improvement in performance, energy efficiency and compactness. 

Many existing works have explored hardware implementations of DCNNs using GPUs \cite{krizhevsky2012imagenet} and FPGAs \cite{rahman2016efficient,motamedi2016design}.
Nevertheless, more efficient design is required to resolve the conflict between the resource-hungry DCNNs and the resource-constrained IoT entities. 
Stochastic Computing (SC), which uses the probability of 1s in a random bit stream to represent a number, has the potential to enable massively parallel and ultra-low footprint hardware based DCNNs \cite{li2017towards,brown2001stochastic1,li2016dscnn,kim2016dynamic,asplos}.
In SC, arithmetic operations like multiplication can be performed using simple logic elements and SC provides better soft error resiliency \cite{li2016joint,alaghi2013survey,brown2001stochastic1,li2016accelerating}.
In this regard, considerable efforts have been invested in the context of designing Artificial Neural Networks (ANNs), Deep Belief Network (DBNs) and DCNNs using SC components \cite{kim2016dynamic,ji2015hardware,sanni2015fpga,li2017towards,li2016dscnn,asplos,listructural}. 

One key challenge is designing accurate nonlinear activation function. 
A small imprecision induced by the convolution and down sampling operations can be significantly amplified by the inaccurate nonlinear activation function, propagated to the subsequent neurons, and further amplified by the activation functions in the following neurons.
Hence, without an accurate activation, the network accuracy can easily decrease to an unacceptable level. 
Besides, while the type of activation has a significant impact on the performance of DCNNs, only two basic Finite State Machine (FSM) based hyperbolic tangent (tanh) activations are designed for ANNs and DBNs in \cite{brown2001stochastic1,kim2016dynamic}, whilst other possible activations have hardly been explored, especially the Rectified Linear Units (ReLUs), which are essential for the state-of-the-art neural networks \cite{he2015delving}. 



In this paper, we first propose three accurate SC-based neuron designs for DCNNs with the popular activation functions that are widely used in the software, i.e., tanh, logistic (or sigmoid), and ReLU. 
The SC configuration parameters are jointly optimized considering the down sampling operation, block size, and connection patterns in order to yield the maximum precision. 
Then we conduct a comprehensive comparison of the aforementioned neuron designs using different activations under different input sizes and stochastic bit stream lengths. 
After constructing the DCNNs using the proposed neurons, we further evaluate and compare the network performance of DCNNs. 
Experimental results demonstrate that proposed SC based DCNNs have much smaller area and power consumption than the binary ASIC DCNNs, with up to 61X, 151X, 2X improvement in terms of area, power, and energy, respectively, at the cost of $0.0001 - 0.03$ accuracy degradation. 
Moreover, the SC approach achieves up to 21X and 41X of the area, 41X and 72X of the power, and 198200X and 96443X of the energy, compared with CPU and GPU approaches, respectively, while the test error is increased by less than 3.07\%.
ReLU activation is suggested for future SC based DCNNs considering its superior accuracy, area, and energy performance under a small bit stream length. 


The remainder of the paper is organized as follows. 
Section II reviews related work.
DCNN architecture and related SC components are given in Section III. 
Section IV presents the three activation function designs in a neuron cell using SC. 
Section V reports the experimental results, and this paper is concluded in Section VI.

\section{Related Work}

\subsection{Activation Function Studies}

The hyperbolic tangent has generally shown to provide more robust performance than logistic function in DBNs \cite{maas2013rectifier}. 
However, both suffer from the vanishing gradient problem, resulting in a slowed training process or convergence to a poor local minimum \cite{maas2013rectifier}. 
ReLUs do not have this problem, since an activated unit gives a constant
gradient of 1. 
Recently, the parametric ReLU proposed by K. He et al. was reported to surpass human-level performance on the ImageNet large scale visual recognition challenge \cite{he2015delving}. 

\subsection{Hardware-Based DCNN Studies}


In order to exploit the parallelism and reduce the area, power, and energy, 
many existing hardware-based DCNNs have come into existence, including GP-GPUs based DCNNs \cite{ciresan2011flexible} and FPGA-based DCNNs \cite{motamedi2016design, rahman2016efficient}.
GP-GPU was the most commonly used platform for accelerating DBNs around 2011 \cite{ciresan2011flexible}.  
As for FPGA, M. Motamedi et al. developed an FPGA-based accelerator to meet performance
and energy-efficiency constraints of DCNNs \cite{motamedi2016design}.
Recently, SC becomes a very attractive candidate for implementing ANNs and DBNs. 
Y. Ji et al. applied SC to a radial basis function ANN and significantly reduced the required hardware in \cite{ji2015hardware}, and hardware-oriented optimization for SC based DCNN was developed in \cite{li2016dscnn}. 
K. Kim et al. proposed hardware-based DBN using SC components, in which a SC based neuron cell was designed and optimized \cite{kim2016dynamic}.
A multiplier SC neuron and a structure optimization method were proposed in \cite{li2017towards} for DCNN. 
A. Ren et al. developed a DCNN architecture with weight storage optimization and a novel max pooling design in the SC domain \cite{asplos}. 
Besides, Z. Li et al. explored eight different neuron implementations in DCNNs using SC \cite{listructural}. 

However, no existing works have explored the ReLU and logistic activations (popular activation choices in software) for hardware based DCNNs using SC. 

\section{Overview of Hardware-Based DCNN}


\subsection{General Architecture of DCNNs}

\begin{figure}[b]
	\centering
	\includegraphics[width=1\columnwidth]{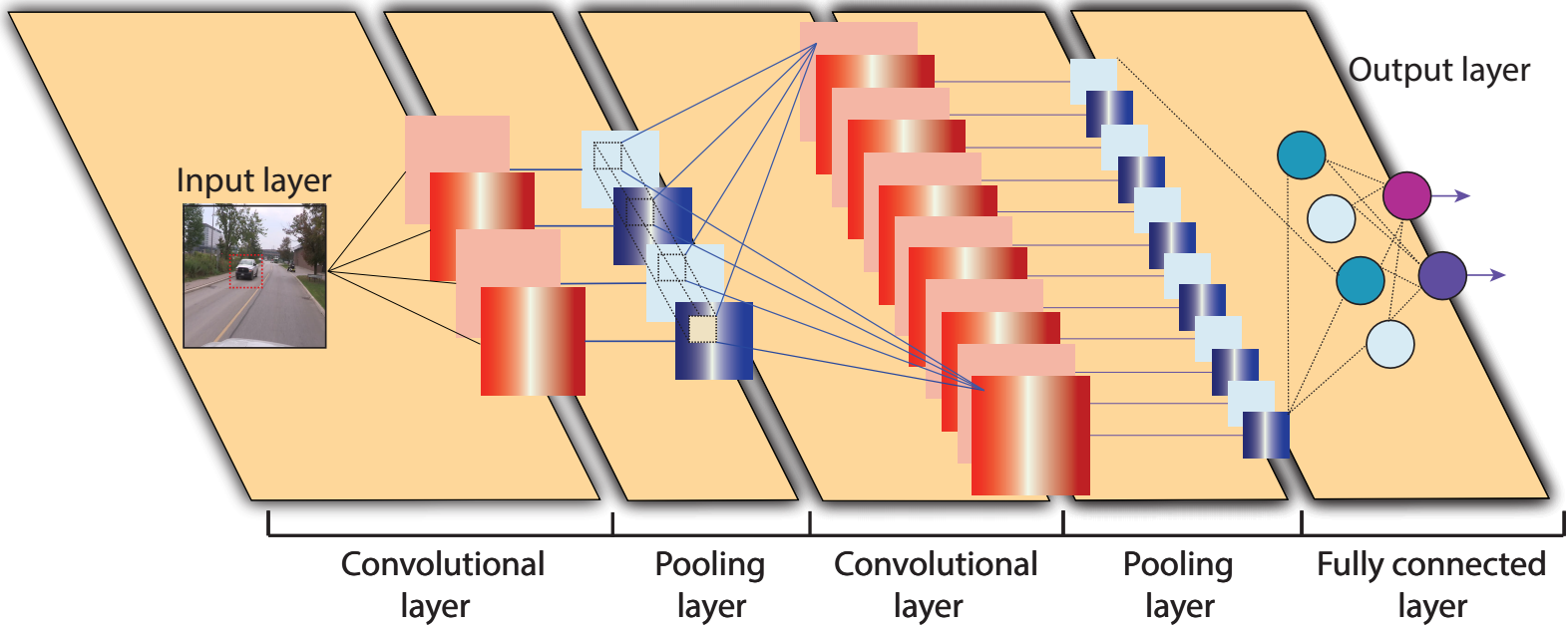}
	\vskip -0.8em
	\caption{A general DCNN architecture. }
	\label{fig:overview}
\end{figure}

Figure \ref{fig:overview} shows a general DCNN architecture that is composed of a stack of convolutional layers, pooling layers, and fully connected layers. 

In each convolutional layer, common patterns in local regions of inputs are extracted by convolving a filter over the inputs. 
The convolution result is stored in a feature map as a measure of how well the filter matches each portion of the inputs. 
After convolution, a subsampling step is performed by the pooling layer to aggregate statistics of these features, reduce the dimensions of data and mitigate over-fitting issues. 
A nonlinear activation function is applied to generate the output of the layer. 
The stack of convolutional and pooling layers is followed by fully connected layers, which further aggregate the local information learned in the convolutional and pooling layers for class discrimination. 

By alternating the topologies of convolutional and pooling layers, powerful DCNNs can be built for specific applications, such as LeNet \cite{lecun1998gradient}, AlexNet \cite{krizhevsky2012imagenet}, and GoogLeNet \cite{szegedy2015going}.
With no loss of generality, we use LeNet-5 (i.e., the fifth generation of LeNet for digits recognition) in our discussion and experiments throughout the paper, and the proposed design methodology can accommodate other DCNNs as well.

\begin{figure}[t]
	\centering
	\includegraphics[width=1\columnwidth]{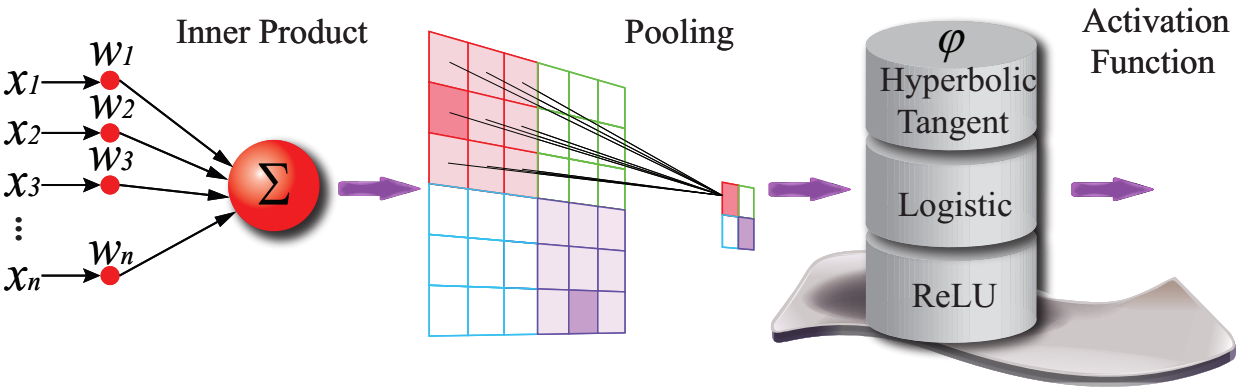}
	\vskip -0.8em
	\caption{A hardware-based neuron cell in DCNN.}
	\label{fig:neuron}
\end{figure}

\subsection{Hardware-Based Neuron Cell}

As the basic building block in DCNNs, a neuron performs three fundamental operations, i.e., convolution, pooling, and activation, as shown in Figure \ref{fig:neuron}. 

The convolution operation calculates the inner products of input ($x_i's$) and weights ($w_i's$), and the pooling operation performs sub-sampling for the inner product results, i.e., generating one result out of several inner products. 
There are two conventional choices for pooling: max and average. 
The former chooses the largest element in each pooling region, whereas the latter calculates the arithmetic mean. 
An activation function is applied before the output, with tanh $f(x)=tanh(x)$, logistic function  $f(x)=(1+e^{-x})^{-1}$, and ReLU $f(x)=max(0,x)$ being popular choices. 

From a hardware design perspective, convolution and pooling can be implemented efficiently using basic building blocks (described in Section \ref{sc_nd}) to achieve accurate results. 
The nonlinear activation function, on the other hand, is usually implemented using Look-Up Tables (LUTs), requiring large memories to achieve adequate precision.
Moreover, without careful design and optimization, the nonlinear activation function can result in serious accuracy degradation as it directly affects the final output accuracy of the neuron cell and the imprecision can be amplified by the subsequent calculations and activations. 
Hence, it is imperative to design and optimize the activation function to achieve sufficient accuracy.

\section{Proposed Hardware-Driven Nonlinear Activation for DCNNs}

\subsection{Stochastic Computing for Neuron Design}\label{sc_nd}

Stochastic computing (SC) is an encoding scheme that represents a numeric
value $x\in [0,1]$ using a bit stream $X$, in which the probability of ones is $x$ \cite{brown2001stochastic1}. 
For instance, the bit stream \enquote{01000} contains a single one in a five-bit stream, thus it represents $ x=P(X=1)=0.2 $. 
In addition to this unipolar encoding format, another commonly used format is 
bipolar format, where a numeric value $x\in [-1,1]$ is processed by $P(X=1)=\frac{x+1}{2}$. 
Using bipolar format, the previous example $0.2$ could be represented by \enquote{10110}. 
Next, we present the fundamental operations of a neuron using SC components. 

\begin{figure}[t]
	\centering
	\includegraphics[width=0.85\columnwidth]{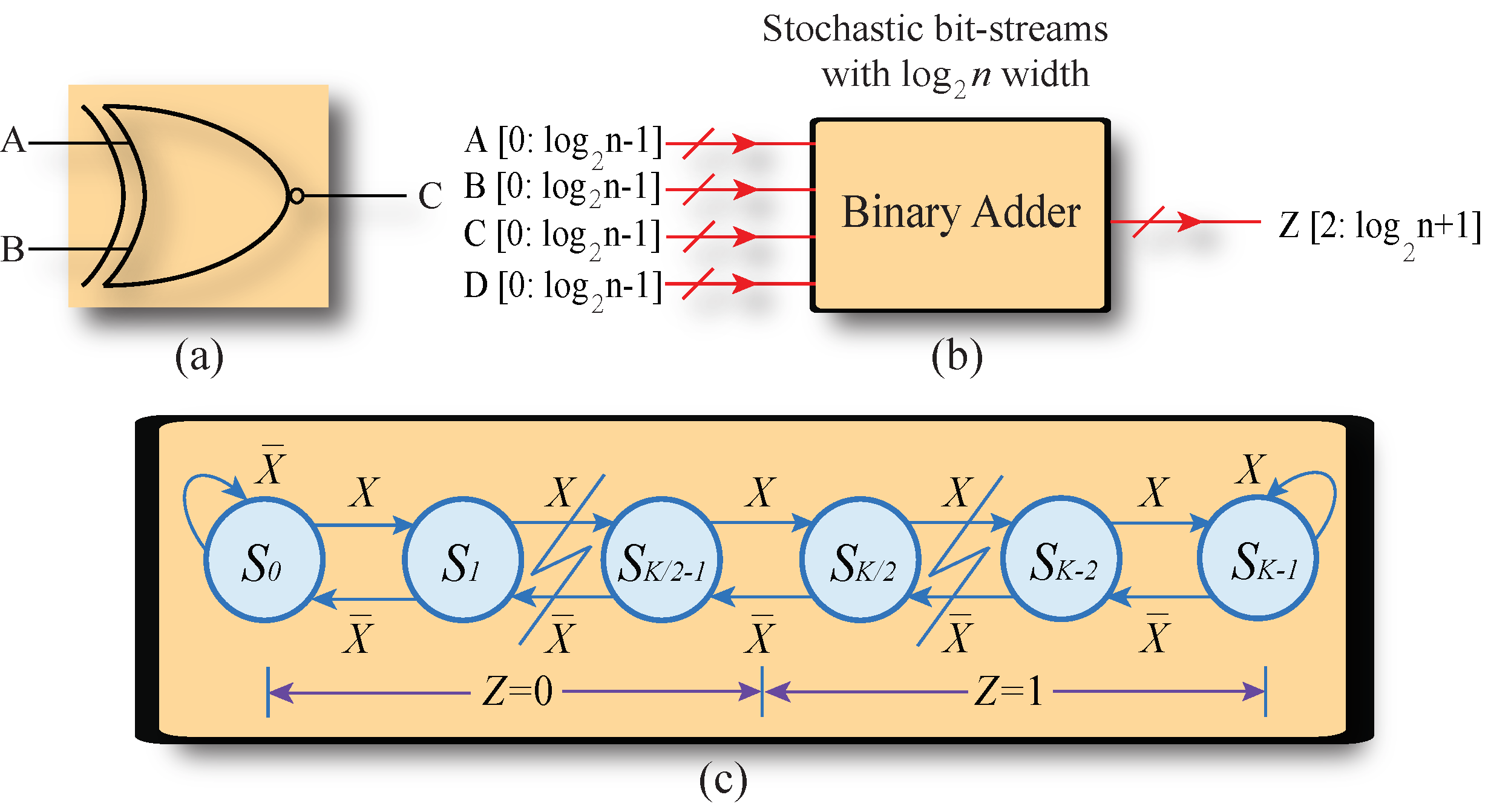}
	\vskip -0.8em
	\caption{Stochastic computing for neuron design: (a) XNOR gate for bipolar multiplication, (b) binary adder for average pooling, and (c) FSM-based tanh for stochastic inputs. }
	\label{fig:sc}
\end{figure}

\textbf{Convolution}. 
Convolution performs multiplication and addition. 
An XNOR gate, as shown in Figure \ref{fig:sc} (a), is used for multiplication with bipolar stochastic encoding as  $c=2P(C=1)-1=2(P(A=1)P(B=1)+P(A=0)P(B=0))-1=(2P(A=1)-1)(2P(B=1)-1)=a \times b $.
Multiplexers (MUXes) can perform bipolar SC addition \cite{brown2001stochastic1}, which randomly selects one input as output. 
When input size is large, summation using MUXes incurs significant accuracy loss since only one input is selected at a time and all the other inputs are not used. 
To achieve a good trade-off between accuracy and hardware cost in terms of area, power, and energy, we adopt the Approximate Parallel Counter (APC) developed in \cite{kim2015approximate} for addition instead of MUXes. 

\textbf{Pooling}. 
In the SC domain, average pooling can be implemented efficiently with a simple MUX for stochastic inputs. 
However, as the outputs of APCs (i.e., inputs of pooling) are binary, MUX cannot be applied here. 
Instead, we use a binary adder to calculate the sum and remove the last 2 bits of the sum as a division by 4 operation, as illustrated in Figure \ref{fig:sc} (b). 
Note that 4-to-1 average pooling is used in the LeNet-5, whereas max pooling is used in other DCNNs, such as AlexNet \cite{krizhevsky2012imagenet} and GoogLeNet \cite{szegedy2015going}. 
The SC-based max pooling design is developed in \cite{asplos}. 

\textbf{Activation}. 
Nonlinear activation is the key operation that enhances the representation capability of a neuron. 
There are many types of activation functions that have been explored in software DCNNs.
Nevertheless, due to the design difficulty, only tanh is designed in the SC domain (e.g., an FSM-based tanh function $Stanh(\cdot)$ proposed in \cite{brown2001stochastic1}, as shown in Figure \ref{fig:sc} (c)), 
and there is a serious lack of studies on other activation functions for SC-based DCNNs.
For the convenience of discussions, we use the naming conventions in Table \ref{tbl_name}. 

\subsection{Proposed Neuron Design and Nonlinear Activation}

We found that directly applying SC to DCNNs leads to severe accuracy degradation which is not acceptable in common cases. 
The main reason is that the calculation of each computation block (convolution, pooling, and activation) is inaccurate in the SC domain. 
The overall accuracy of a neuron is affected by number of input streams, bit stream length, hardware configurations, and most importantly the activation functions. 
In this paper, we design and optimize the neuron structures as well as the activations by jointly considering all the aforementioned factors in order to reach the accuracy level that can be achieved by fixed point binary arithmetic. 

\begin{table}[b]\scriptsize
	\centering
	\caption{Naming Conventions in a Stochastic Computing Based Neuron}
	\label{tbl_name}
	\vskip -0.8em
		\resizebox{1\columnwidth}{!}{
	\begin{tabular}{c|l}
		\hline
		$m$ & the length of bipolar bit stream \\ \hline
		$q$ & $q$-to-$1$ average pooling \\ \hline
		\multirow{2}{*}{$n$} & 	input size: the number of input bit streams (or the number of \\
		& input and weight pairs) in each convolution block \\ \hline
		\multirow{2}{*}{$x_i^j$} & the $i$-th input bit of the $j$-th convolution block \\ 
		& $\big(i\in [0,n-1]$, $j\in [0,q-1]\big)$ \\ \hline
		\multirow{2}{*}{$w_i^j$} & the $i$-th weight bit of the $j$-th convolution block \\ 
		& $\big(i\in [0,n-1]$, $j\in [0,q-1]\big)$ \\ \hline
		\multirow{2}{*}{$\phi_j$} & the set of all the input bits of the $j$-th convolution block\\ 
		& 	$\big(\phi_j=\{x_i^j| j\in[0,q-1], \forall i\in [0,n-1]\}\big)$ \\ \hline
		\multirow{2}{*}{$\psi_j$} & the set of all the weight bits of the $j$-th convolution block\\ 
		& 	$\big(\psi_j=\{w_i^j| j\in[0,q-1], \forall i\in [0,n-1]\}\big)$ \\ \hline	
		\multirow{2}{*}{$t_j$} & the sum calculated by the $j$-th convolution block,  \\
		& which is a $log_2 n$-bit binary number \\ \hline	
		$e$ & the number of states in the state machine \\ \hline	
		$z_k$ & the $k$-th stochastic output bit for the SC-based hyperbolic tangent neuron \\ \hline		
	\end{tabular}
		}
\end{table}

To be more specific, we adopt bipolar encoding scheme and average pooling. 
The proposed neuron designs use XNOR gates and APCs for addition and multiplication (as convolution operation), respectively.
Average pooling is implemented using a binary adder (as shown in Figure \ref{fig:sc} (b)). 
The nonlinear activations are designed for binary inputs since the output of APC and the pooling result are binary. 
The output of the proposed neuron, on the other hand, must be a stochastic bit stream, which will be fed into the neurons in the subsequent layers. 
The proposed neurons using hyperbolic tangent, stochastic logistic and ReLU activations are as follows:

\textbf{SC-tanh: SC-Based Hyperbolic Tangent Neuron.}
The authors in \cite{kim2016dynamic} adopted a saturated up/down counter to implement a binary hyperbolic tangent activation function. 
Nevertheless, this activation function is designed for DBNs and cannot be applied directly for DCNNs. 
Algorithm \ref{algo_tanh} presents the proposed SC-based hyperbolic tangent neuron design \textit{SC-tanh}($\cdot$) for DCNNs, where step 5-9 correspond to the inner product calculation using XNOR gates and APCs, step 11 is the average pooling with a binary adder, and step 12-18 are the tanh activation part, which is implemented with a saturated counter. 
The connections between two layers are based on the filters that execute convolution operations only for portions of the inputs (i.e., receptive fields), which greatly reduce the connections between consecutive layers. 
The counted binary number generated by APC is taken by the saturated up/down counter which represents the amount of increase and decrease. 
By setting half of the states' output to 1 and the rest to 0 (i.e., setting boundary state to $S_{bound}=\frac{e}{2}$), this neuron design accurately imitates the hyperbolic tangent function.

\begin{algorithm}[t]\scriptsize		
\SetKwData{S}{$S$}\SetKwData{Smax}{$S_{max}$}\SetKwData{Shalf}{$S_{bound}$}\SetKwData{temp}{$temp$}\SetKwData{outi}{$out_{i}$}
\SetKwInOut{Input}{input}\SetKwInOut{Output}{output}

\Input{$\phi, \psi$ are the input bit streams}
\Input{$q$ indicates the $q$-to-$1$ average pooling}
\Input{$e$ is internal FSM state number}
\Output{$z_k$ is the $k$-th stochastic output bit for the SC-tanh neuron }
\Smax$\leftarrow$\ {$e-1$}  \tcc*[r]{max state}
\Shalf$\leftarrow$\ {$e/2$} \tcc*[r]{boundary state}
\S$\leftarrow$ \Shalf \tcc*[r]{current state}
\For{$k\leftarrow 1$ \KwTo $m$}{
	\tcc{processing each convolution block}
	\For{$j\leftarrow 1$ \KwTo $q$}{
		\tcc{inner product calculation}
		\For{$i\leftarrow 1$ \KwTo $n$}{
			$p_i^j=x_i^j \odot w_i^j$ \tcc*[r]{XNOR multiplication}
		}
		$t_j = 2\cdot \sum_{i=0}^n p_i^j - n$ \tcc*[r]{APC addition}
	}
	\tcc{average pooling and tanh activation}
	\S$\leftarrow$\ {\S $+ \frac{\sum_{j=1}^{q} t_j}{q}$} \\
	\uIf(\tcc*[f]{saturated counter}){\S$<0$}{
		\S $\leftarrow$ 0\;
	}
	\uElseIf{\S$>$\Smax}{
		\S $\leftarrow$ \Smax 
	}
	\eIf(\tcc*[f]{output logic}){\S$>$\Shalf}{
		$z_k \leftarrow 1$
	}{
	$z_k \leftarrow 0$
}
}
\caption{Proposed SC-tanh ($\phi,\psi,q,e$)}\label{algo_tanh}
\end{algorithm}

\textbf{SC-logistic: Proposed SC-Based Logistic Neuron.}
There are two important differences between the hyperbolic tangent neuron and the logistic neuron: 
(i) unlike the above-mentioned hyperbolic tangent neuron, the output of which is in the range of [-1,1], the logistic neuron always outputs non-negative number (i.e., within [0,1]), 
and (ii) the $y$-value of the logistic's midpoint is $0.5$ instead of $0$ in hyperbolic tangent.

In order to tackle the $1$-st difference, we introduce a history shift register array $H[0:\alpha-1]$ using $\alpha$ shift registers to record the last $\alpha$ bits of the output stochastic bit streams. 
A shadow counter is used to calculate the sum of the stochastic bits in the history shift register array, which is denoted by $\delta$, i.e., $\delta = \sum_{i=0}^{\alpha-1}H[i]$.
Hence, the proposed SC-logistic keeps tracking the last $\alpha$ output bits and predicts the sign of the current value based on the sum calculated by the shadow counter. 
To be more specific, if the sum is less than half of the maximum achievable sum $\delta < \frac{\alpha}{2}$, the current value is predicted to be negative. Otherwise, it is predicted as positive (note that value $0$ is half 1s and half 0s in bipolar format). 
As any negative output raises an error, the proposed activation mitigates such errors by outputting 1 (as compensation) whenever the predicted current value is negative. 

As for the $2$-nd difference, we need to move the $y$-value of the logistic's midpoint to $0.5$. 
The probability of 1s in a stochastic bit stream $X$ that represents the value $x=0.5$ is $P(X=0.5)=\frac{3}{4}$. 
Hence, we design the logic of activation such that $\frac{1}{4}$ of the states output $0$s, whereas the other $\frac{3}{4}$ portion output $1$s. 
This is realized by setting the boundary state to $S_{bound}=\frac{e}{4}$, where $e$ represents the internal FSM state number (in the saturated counter). 
Note that the boundary value in hyperbolic tangent is $S_{bound}=\frac{e}{2}$ since the midpoint of $y$-value of tanh is $0$ and $0$ is represented by a stochastic stream where half of the bits is $0$ and the other half is $1$ (i.e., $P(X=0)=\frac{1}{2}$). 
Algorithm \ref{algo_losigtic} provides the pseudo code of the proposed SC-logistic neuron.




\begin{algorithm}[]\scriptsize		
	\SetKwData{S}{$S$}\SetKwData{Smax}{$S_{max}$}\SetKwData{Shalf}{$S_{bound}$}\SetKwData{temp}{$temp$}\SetKwData{outi}{$out_{i}$}
	
	\SetKwInOut{Input}{input}\SetKwInOut{Output}{output}
	
	\Input{$\phi, \psi$ are the input bit streams}
	\Input{$\alpha$ is the number of registers in the temporary array}	
	\Input{$\beta$ is the configuration bit. 1:SC-logistic, 0:SC-ReLU}
	\Input{$q$ indicates the $q$-to-$1$ average pooling}
	\Input{$e$ is internal FSM state number}
	\Output{$z_k$ is the $k$-th stochastic output bit for the SC-logistic/ReLU neuron}
	
	\Smax$\leftarrow$\ {$e-1$}  \tcc*[r]{max state}
	
	\eIf(\tcc*[f]{boundary state configuration}){$\beta == 1$}{
		\Shalf$\leftarrow$\ {$e/4$} \tcc*[r]{SC-logistic}
	}{
	\Shalf$\leftarrow$\ {$e/2$} \tcc*[r]{SC-ReLU}
}
\S$\leftarrow$ \Shalf \tcc*[r]{current state}
$r \leftarrow \alpha-1$ \tcc*[r]{$r$ is an iterator}
$H[0:\alpha-1] \leftarrow 0$ \tcc*[r]{initialize history array}
$\delta \leftarrow 0$ \tcc*[r]{initialize shadow counter}

\For{$k\leftarrow 1$ \KwTo $m$}{
	
	\eIf{$\delta < \frac{\alpha}{2}$}{
		$z_k \leftarrow 1$ \tcc*{negative value compensation}
	}{
	\For{$j\leftarrow 1$ \KwTo $q$}{
		\For{$i\leftarrow 1$ \KwTo $n$}{
			$p_i^j=x_i^j \odot w_i^j$ \tcc*[r]{XNOR multiplication}
		}
		$t_j = 2\cdot \sum_{i=0}^n p_i^j - n$ \tcc*[r]{APC addition}
	}
	\S$\leftarrow$\ {\S $+ \frac{\sum_{j=1}^{q} t_j}{q}$} \\
	\uIf(\tcc*[f]{saturated counter}){\S $<0$}{\S $\leftarrow 0$}	
	\ElseIf{S $>$ \Smax}{\S $\leftarrow$ \Smax}
	\eIf(\tcc*[f]{output logic}){\S $>$ \Shalf}{ $z_k \leftarrow 1$}{$z_k \leftarrow 0$}		
}

\While{$r\geq 1$}{
	$H[r] \leftarrow H[r-1]$\tcc*[r]{update the history array}
	$r \leftarrow r - 1$	
}
$H[0] \leftarrow z_k$\\
$\delta \leftarrow \sum_{r=0}^{\alpha-1}H[r]$\tcc*[r]{update the shadow counter}
$r \leftarrow \alpha -1 $
}

\caption{Proposed SC-logistic/ReLU ($\phi,\psi,\alpha,\beta,q,e$)}\label{algo_losigtic}
\end{algorithm}

\textbf{SC-ReLU: Proposed SC-Based ReLU Neuron.}
According to \cite{lecun2015deep}, the ReLU activation $f(x)=max(0,x)$ becomes the most popular activation function in 2015. 
We apply the same design concept as in the SC-tanh neuron with the following modifications:
First, we introduce a history shift register array $H[0:\alpha-1]$ and its shadow counter $\delta$ to predict the sign of the current value, and compensate the negative value in the same way SC-logistic neuron does. 
Second, unlike the SC-logistic neuron that changes boundary state to $S_{bound}=\frac{e}{4}$ such that the entire waveform is moved up to $y=0.5$, the SC-ReLU is centered at (0,0) so the boundary state needs to be kept as $S_{bound}=\frac{e}{2}$. 
Noticing the similarities between the SC-logistic and SC-ReLU, we introduce a configuration bit $\beta$ and combine these two neurons, and Algorithm \ref{algo_losigtic} provides the pseudo code of the proposed SC-logistic and SC-ReLU neuron.

\begin{figure}[b]
	\centering
	\includegraphics[width=0.98\columnwidth]{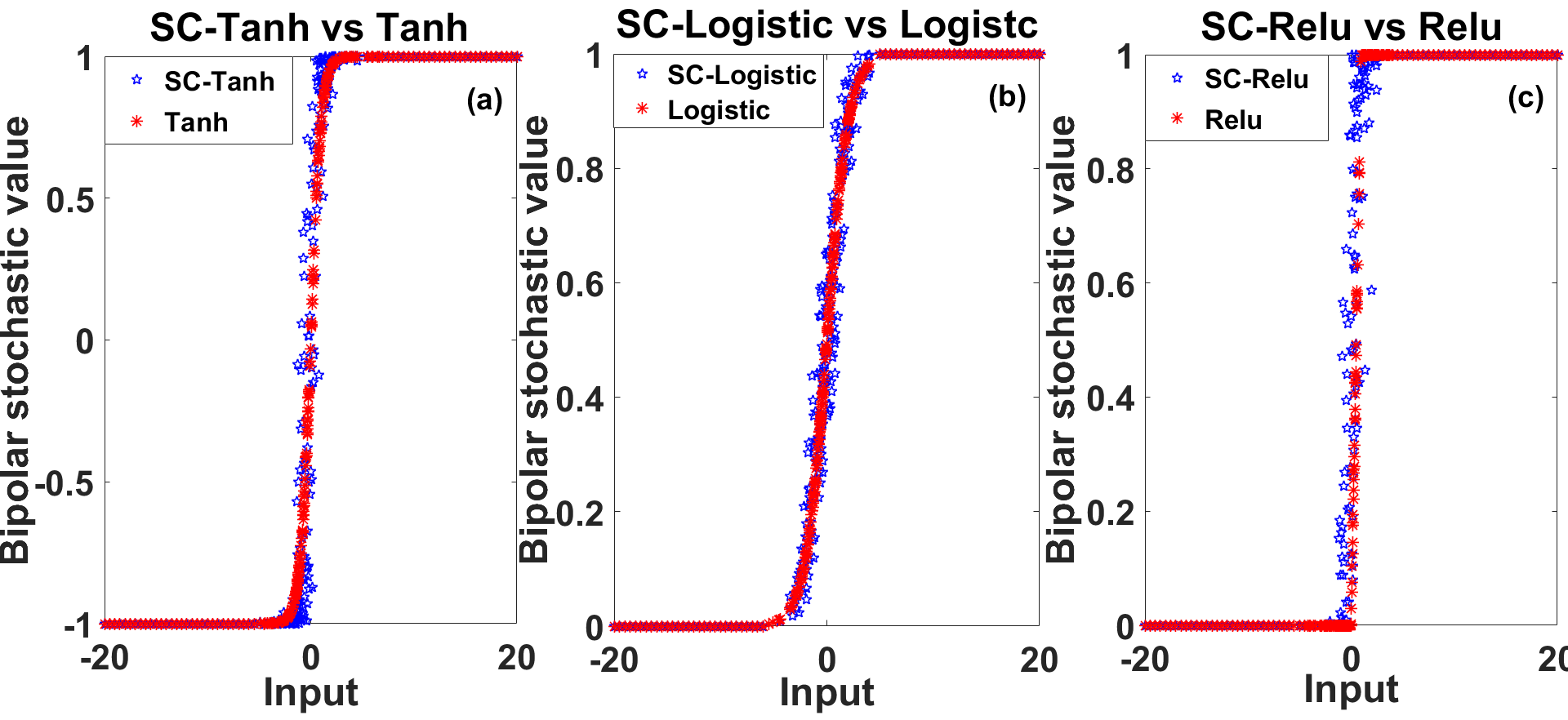}
	\vskip -0.8em
	\caption{ The result comparison between the proposed SC neuron (bit stream $m=1024$) and the corresponding original software neuron: (a) SC-tanh vs Tanh, (b) SC-logistic vs Logistic, and (c) SC-ReLU vs ReLU.}
	\label{fig:neuron_comp}
\end{figure}

Note that in the proposed neurons, the FSM state number $e$ is generated using a simple binary search algorithm under different input sizes to yield the highest precision. 
Unlike the software implementation that has limited allowable degree of parallelism and high coordination overheads, the proposed specific hardware based neurons can execute the commands in Algorithm \ref{algo_tanh} and \ref{algo_losigtic} fully parallelly. 
Figure \ref{fig:neuron_comp} (a), (b), and (c) show that the proposed SC-tanh, SC-logistic, and SC-ReLU neurons and their corresponding software results are almost identical to each other. 
Detailed error analysis is provided in Section \ref{sec:exp}.

\section{Experimental Results} \label{sec:exp}

We now present (i) performance evaluation of the proposed SC neurons, (ii) comparison with binary ASIC neurons, and (iii) DCNN performance evaluation and comparison. 
The neurons and DCNNs are synthesized in Synopsys Design Compiler with the 45nm Nangate Library \cite{nangate} using Verilog.

\subsection{Performance Evaluation and Comparison among the Proposed Neuron Designs}

For each neuron, the accuracy is dependent on (i) the bit stream length $m$, and (ii) the input size $n$. 
Longer bit stream length yields higher accuracy, and the precision can be leveraged by adjusting the bit stream length without hardware modification.
On the other hand, the input size $n$, which is determined by the DCNN topology, affects both the accuracy of the neuron as well as the hardware footprint. 

\begin{figure*}
	\centering
	\includegraphics[width=1\textwidth]{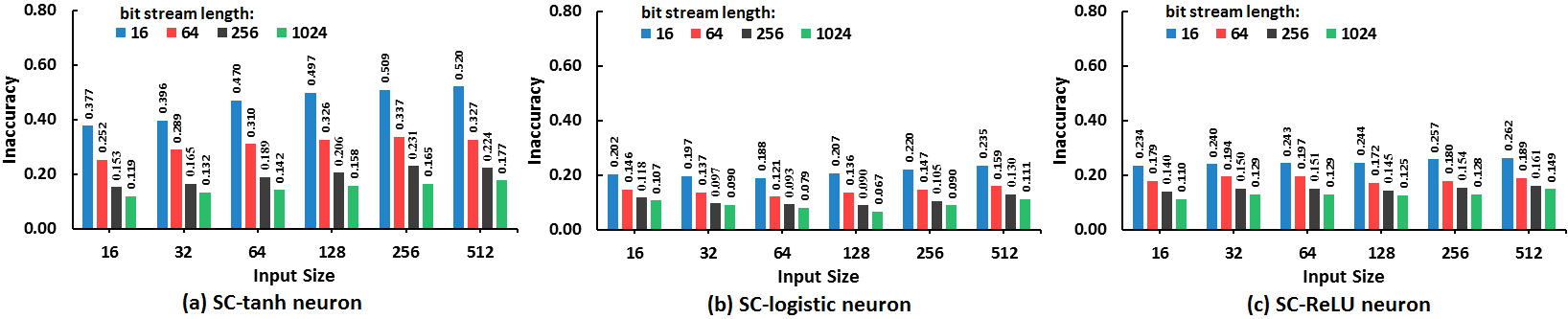}
	\vskip -0.8em
	\caption{Input size versus absolute inaccuracy under different bit stream lengths for (a) SC-tanh neuron, (b) SC-logistic neuron, and (c) SC-ReLU neuron.}
	\label{fig:neuron_accuracy}
\end{figure*}

\begin{figure*}
	\centering
	\includegraphics[width=1\textwidth]{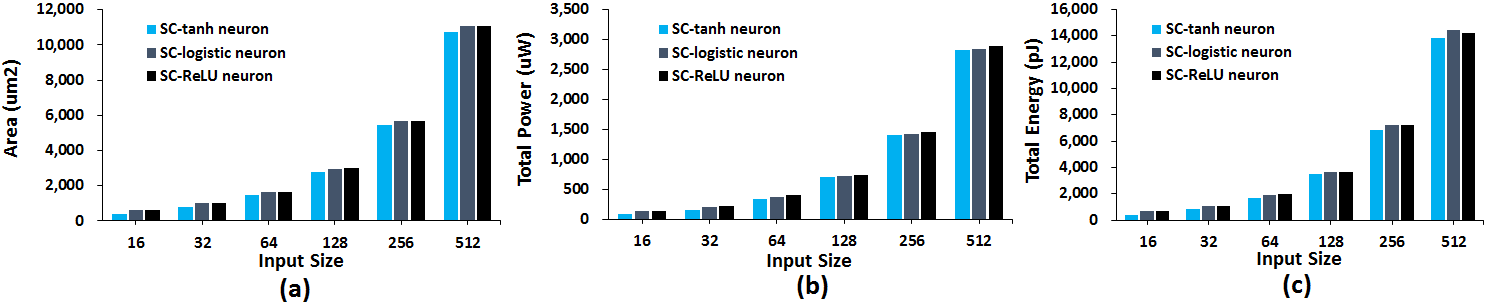}
	\vskip -0.8em
	\caption{Input size versus (a) area, (b) total power, and (c) total energy for the neuron designs using tanh, logistic and ReLU activation functions.}
	\label{fig:neuron_cost}
\end{figure*}

Thus, considering the aforementioned factors, we evaluate the inaccuracy of the proposed SC-tanh, SC-logistic and SC-ReLU neurons under a wide range of bit stream lengths and input sizes, as shown in Figure \ref{fig:neuron_accuracy} (a), (b) and (c), respectively. 
The corresponding hardware costs of the proposed SC-tanh, SC-logistic and SC-ReLU neurons are shown in Figure \ref{fig:neuron_cost} (a), (b) and (c), respectively. 
One can observe that the precisions of SC-logistic and SC-ReLU neurons consistently outperform SC-tanh neurons under different combinations of input size and bit stream length, whereas the area, power and energy of SC-tanh neurons are slightly lower than the other two neurons. 
Moreover, SC-logistic and SC-ReLU neurons have better scalability in terms of accuracy than the SC-tanh neuron. 
Note that the inaccuracy here is calculated by comparing with the software neuron results. 
A lower inaccuracy only indicates that the hardware neuron with this type of activation is closer to its software version, but this does not indicate that a DCNN implemented with this neuron can yield a lower test error.

\begin{table}[b]
	\centering
	\caption{Neuron Cell Performance Comparison with 8 Bit Fixed Point Binary Implementation when $n=25$ and $m=1024$}
	\label{tbl_neuron}
	\vskip -0.8em
	\resizebox{1\columnwidth}{!}{
		\begin{tabular}{c|rrr|rrr|rrr}
			\hline
			& \multicolumn{3}{c|}{8 bit binary ASIC} &	\multicolumn{3}{|c|}{proposed SC neuron} & \multicolumn{3}{|c}{improvement} \\ \cline{2-10}
			& tanh & logistic & ReLU & tanh & logistic & ReLU & tanh & logistic & ReLU \\\hline
			%
			
			power & \multirow{2}{*}{34760} & \multirow{2}{*}{34760} & \multirow{2}{*}{32176} & \multirow{2}{*}{173} & \multirow{2}{*}{223} & \multirow{2}{*}{231} & \multirow{2}{*}{201X} & \multirow{2}{*}{156X} & \multirow{2}{*}{139X} \\
			($\mu W$) & & & & & & & & & \\\hline	 	 	 	 	 	 	 	 
			
			energy & \multirow{2}{*}{128267} & \multirow{2}{*}{128267} & \multirow{2}{*}{94346} & \multirow{2}{*}{858} & \multirow{2}{*}{1130} & \multirow{2}{*}{1147} & \multirow{2}{*}{149X} & \multirow{2}{*}{114X} & \multirow{2}{*}{82X} \\
			($fJ$) & & & & & & & & & \\\hline	 
			
			area & \multirow{2}{*}{42319} & \multirow{2}{*}{42319} & \multirow{2}{*}{35506} & \multirow{2}{*}{699} & \multirow{2}{*}{916} & \multirow{2}{*}{916} & \multirow{2}{*}{61X} & \multirow{2}{*}{46X} & \multirow{2}{*}{39X} \\
			($\mu m^2$) & & & & & & & & & \\\hline	 	
			
		\end{tabular}
	}
\end{table}	

\subsection{Comparison with Binary ASIC Neurons}

We further compare the proposed SC-based neurons with the binary ASIC hardware neurons. 
The input size is set to 25 since most neurons in LeNet-5 DCNN are connected to $5\times 5$ receptive fields. 
The binary nonlinear activations logistic and ReLU are implemented using LUTs, whereas binary ReLU is built with a comparator and a MUX.
Clearly, the number of bits in fixed-point numbers affects both the hardware cost and the accuracy. 
To make the comparison fair, we use the minimum fixed point (8 bit) that yields a DCNN network accuracy that is almost identical to the software DCNN (with $< 0.0003$ difference in network test error). 
Table \ref{tbl_neuron} shows the neuron cell performance comparison between the proposed SC neurons and the 8 bit binary ASIC neurons. 
Compared with binary ASIC neurons, the proposed SC neurons achieve up to 201X, 149X, and 61X improvement in terms of power, energy, and area, respectively, indicating significant hardware savings. 

\begin{table*}
	\centering
	\caption{Comparison among Software DCNN, Binary ASIC DCNN, and Various SC Based DCNN Designs Implementing LeNet 5}
	\label{tbl_network}
	\vskip -0.8em
	\resizebox{0.65\textwidth}{!}{
		\begin{tabular}{cccc cccccc c}
			\hline
			activation &  approach & bit stream & valid. error & test error  & area (mm2) & power ($W$) & energy ($\mu J$) \\  \hline			
			
\multirow{8}{*}{tanh} & \multirow{5}{*}{SC} & 1024 &1.74\% & 1.41\% &  \multirow{5}{*}{12.5} & \multirow{5}{*}{3.1} & 15.8 \\ 
&  & 512 &1.57\% & 1.65\%  & & & 7.9  \\ 
&  & 256 &1.71\% & 1.61\%  & & & 3.9 \\ 
&  & 128 &1.84\% & 2.13\%  & & & 2.0  \\ 
&  & 64 &2.37\% & 2.34\%  & & & 1.0 \\ \cline{2-8}
& binary & - & 1.42\% & 1.34\%  & 769.3 & 470.0 & 2.0 \\
& CPU & - &1.41\% & 1.34\%  &263 &130.0 &198200  \\ 
& GPU & - &1.41\% & 1.34\%  &520 &225.0 &96443  \\ \hline

\multirow{8}{*}{logistic} & \multirow{5}{*}{SC} & 1024 &3.98\% & 4.49\% &  \multirow{5}{*}{15.8} & \multirow{5}{*}{3.9} & 20.1 \\ 
&  & 512 &4.35\% & 4.70\%  & & & 10.0  \\ 
&  & 256 &4.24\% & 4.34\%  & & & 5.0 \\ 
&  & 128 &5.23\% & 5.58\%  & & & 2.5 \\ 
&  & 64 &6.30\% & 6.06\%  & & & 1.3 \\ \cline{2-8}
& binary & - & 2.88\% & 3.01\%  & 769.3 & 585.7 & 2.4 \\
& CPU & - &2.87\% & 2.99\%  &263 &130.0 &198200  \\ 
& GPU & - &2.87\% & 2.99\%  &520 &225.0 &96443  \\ \hline			
   
\multirow{8}{*}{ReLU} & \multirow{5}{*}{SC} & 1024 &1.69\% & 1.69\% &  \multirow{5}{*}{15.8} & \multirow{5}{*}{3.9} & 20.3 \\ 
&  & 512 &1.67\% & 1.69\%  & & & 10.1  \\ 
&  & 256 &1.67\% & 1.63\%  & & & 5.1 \\ 
&  & 128 &1.65\% & 1.67\%  & & & 2.5 \\ 
&  & 64 &1.67\% & 1.63\%  & & & 1.3 \\ \cline{2-8}
& binary & - & 1.65\% & 1.65\%  & 664.9 & 557.5 & 1.8 \\
& CPU & - &1.64\% & 1.64\%  &263 &130.0 &198200  \\ 
& GPU & - &1.64\% & 1.64\%  &520 &225.0 &96443  \\ \hline

		\end{tabular}
	}
\end{table*}

\subsection{DCNN Performance Evaluation and Comparison}

To evaluate the network performance, we construct the LeNet-5 DCNNs using the proposed SC neurons as well as the 8 bit binary neurons in a pipelined manner. 
LeNet 5 is a widely-used DCNN structure with a configuration of 784-11520-2880-3200-800-500-10.
The DCNNs are evaluated with the MNIST handwritten digit image dataset \cite{deng2012mnist}, which consists of 60,000 training data and 10,000 testing data.
We apply the same training time in software so as to make a fair comparison among different activations.

Table \ref{tbl_network} concludes the performance of DCNNs using CPU, GPU, binary neurons and the proposed SC neurons. 
The CPU approach uses two Intel Xeon W5580, whereas the GPU approach utilizes NVIDIA Tesla C2075. 
Note that the power for software is estimated using Thermal Design Power (TDP), and the energy is calculated by multiplying the run time and TDP. 
On the other hand, the power, energy, and area for hardware are calculated using the synthesized netlists with Synopsys Design Compiler. 
One can observe that for each type of activation, the proposed SC based DCNNs have much smaller area and power consumption than the corresponding binary DCNNs, with up to 61X, 151X, 2X improvement in terms of area, power, and energy, respectively. 
Note that though the the binary ASIC has a competitive energy performance, it is an ideal pipelined structure. The extremely large area ($>600mm^2$) and power ($>400W$) makes binary ASIC unpractical for implementation. 
Moreover, Table \ref{tbl_network} shows that the proposed SC approach achieves up to 21X and 41X of the area, 41X and 72X of the power, and 198200X and 96443X of the energy, compared with CPU and GPU approaches, respectively, while the error is increased by less than 3.07\%.

Among different activations, with a long bit stream ($m > 128$), SC-tanh is the most accurate. Otherwise ($m\leq 128$), SC-ReLU has the highest precision. 
SC-logistic has the lowest precision due to the following reasons: (i) logistic activation in software has the worst accuracy performance, and (ii) the imprecision of activation (if larger than a certain threshold) amplifies the inaccuracy in DCNNs. 
The bit stream length can be reduced to improve energy performance. 
One important observation is that the proposed SC-ReLU has better scalability than SC-tanh (i.e., with bit stream length decreasing, the accuracy degradation of SC-ReLU is slower than SC-tanh). 
Hence, ReLU activation is suggested for future SC based DCNNs considering its superior accuracy, area, and energy performance under a small bit stream length (e.g., $m=64$). 
Note that the small bit stream length leads to significant improvement in terms of delay and energy performance.

\section{Conclusion}
In this paper we presented three novel SC neurons designs using tanh, logistic, and ReLU nonlinear activations. 
LeNet-5 DCNNs were constructed using the proposed neurons. Experimental results on the MNIST dataset demonstrated that compared to the binary ASIC DCNN, the proposed SC based DCNNs were able to significantly reduce the area, power and energy footprint with a small accuracy degradation. 
ReLU was suggested for future SC based DCNNs implementations.

\bibliographystyle{IEEEtran}
\tiny
\bibliography{sigproc}


%
%
%
%
\end{document}